\documentclass[lettersize,journal]{IEEEtran}
\usepackage{amsmath,amsfonts}
\usepackage{algorithmic}
\usepackage{algorithm}
\usepackage{array}
\usepackage[caption=false,font=normalsize,labelfont=sf,textfont=sf]{subfig}
\usepackage{textcomp}
\usepackage{stfloats}
\usepackage{url}
\usepackage{verbatim}
\usepackage{graphicx}
\usepackage{cite}
\usepackage{multirow}
\usepackage{url}
\usepackage{hyperref}

\begin{document}

\title{LISArD: Learning Image Similarity to Defend Against Gray-box Adversarial Attacks}

\author{Joana C. Costa,
        Tiago Roxo,
        Hugo Proença,~\IEEEmembership{Senior Member,~IEEE,}
        Pedro R. M. Inácio,~\IEEEmembership{Senior Member,~IEEE}
\thanks{Manuscript received February XX, 2025; revised XX XX, 2025. This work was supported in part by the Portuguese Fundação para a Ciência e Tecnologia (FCT)/Ministério da Ciência, Tecnologia e Ensino Superior (MCTES) through National Funds and co-funded by EU funds under Project UIDB/50008/2020; in part by the FCT Doctoral Grant 2020.09847.BD and Grant 2021.04905.BD.}
\thanks{Joana C. Costa, Tiago Roxo, Hugo Proença and Pedro R. M. Inácio are with the Instituto de Telecomunicações, sins-lab, and Department of Computer Science, Universidade da Beira Interior, Portugal (corresponding author e-mail: \href{mailto:joana.cabral.costa@ubi.pt}{joana.cabral.costa@ubi.pt}).}}

\markboth{Submitted to IEEE Transactions on Information Forensics and Security, Vol. XX, XXXX}%
{Costa \MakeLowercase{\textit{et al.}}: LISArD: Learning Image Similarity to Defend Against Gray-box Adversarial Attacks}

\IEEEpubid{0000--0000/00\$00.00~\copyright~2021 IEEE}

\maketitle

\begin{abstract}

State-of-the-art defense mechanisms are typically evaluated in the context of white-box attacks, which is not realistic, as it assumes the attacker can access the gradients of the target network. To protect against this scenario, \textit{Adversarial Training} (AT) and \textit{Adversarial Distillation} (AD) include adversarial examples during the training phase, and \textit{Adversarial Purification} uses a generative model to reconstruct all the images given to the classifier. This paper considers an even more realistic evaluation scenario: \textit{gray-box attacks}, which assume that the attacker knows the architecture and the dataset used to train the target network, but cannot access its gradients. We provide empirical evidence that models are vulnerable to gray-box attacks and propose LISArD, a defense mechanism that does not increase computational and temporal costs but provides robustness against gray-box and white-box attacks without including AT. Our method approximates a cross-correlation matrix, created with the embeddings of perturbed and clean images, to a diagonal matrix while simultaneously conducting classification learning. Our results show that LISArD can effectively protect against gray-box attacks, can be used in multiple architectures, and carries over its resilience to the white-box scenario. Also, state-of-the-art AD models underperform greatly when removing AT and/or moving to gray-box settings, highlighting the lack of robustness from existing approaches to perform in various conditions (aside from white-box settings). All the source code is available at~\url{https://github.com/Joana-Cabral/LISArD}.

\end{abstract}


\section{Introduction}
\label{sec:intro}

\IEEEPARstart{D}{eep} Neural Networks (DNNs) have achieved remarkable performance in multiple areas, such as Medical Imaging~\cite{thirunavukarasu2023large,patricio2023coherent}, Natural Language Processing~\cite{touvron2023llama2openfoundation,costa2022predicting}, and Active Speaker Detection~\cite{roxo2023exploring,roxo2024bias,roxo2024asdnb}. This accomplishment led to the wide adoption of Artificial Intelligence in the daily lives of many people, either in work or leisure scenarios, increasing the attractiveness and susceptibility of DNNs to attackers.
The study of DNN security is still in its early stages. with Szegedy \textit{et al.}~\cite{szegedy2014intriguing} demonstrating, for the first time, that Convolutional Neural Networks (CNNs) fail to generalize and are vulnerable to carefully crafted perturbations (consisting of noise imperceptible to the Human eye) that when added to the original images create the so-called \textit{adversarial examples}.

\begin{figure*}[!t]
    \centering
    \includegraphics[width=0.85\linewidth]{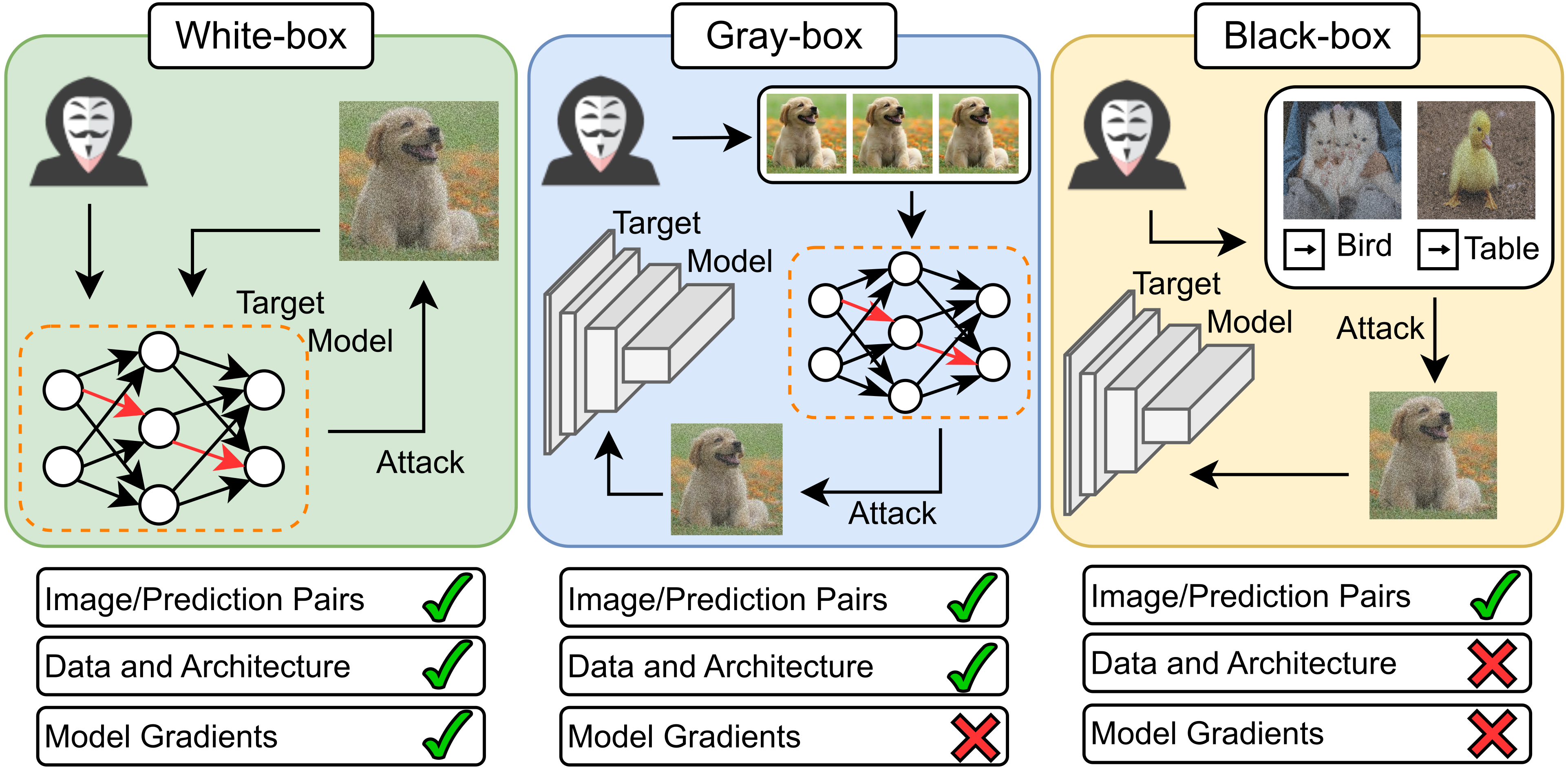}
    \caption{Comparison between the information available to an attacker when considering the different types of attacks. \textit{Image/Predictions Pairs} refers only accessing a set of images given to the model and the respective prediction, \textit{Data and Architecture} refers to knowing the target model architecture and dataset used to train it, and \textit{Model Gradients} refers to controlling the model loss function.}
    \label{fig:attack_types}
\end{figure*}

\textit{Adversarial Distillation}~\cite{papernot2016distillation,goldblum2020adversarially,zhu2021reliable} and \textit{Adversarial Purification}~\cite{yoon2021adversarial,wu2022guided,chen2022densepure} are two of the most studied methods to develop models that are robust against white-box attacks. Although \textit{Adversarial Distillation} can help defend against white-box attacks, it requires including adversarially perturbed images during the training process. \textit{Adversarial Purification} includes a Denoising Diffusion Probabilistic Model (DDPM)~\cite{ho2020denoising} between the inputted image and the target network. This defense mechanism requires the use of a high-parameter model and time to purify each image that is given to the target network. Both previously mentioned defense mechanisms focus on white-box attacks, which are the most explored in the literature and significantly impact the performance of DNNs.

Assuming that the attacker can access the model parameters to generate the perturbed images is unrealistic in many cases. Furthermore, a work by Katzir and Elovici~\cite{katzir2020gradients} found that the ability to defend against white-box attacks comes at the cost of losing the ability to learn. Therefore, this paper proposes a more realistic adversarial scenario that assumes the attacker only knows the network architecture and the dataset used during the training process without accessing model gradients, named the \textbf{gray-box scenario}. Figure~\ref{fig:attack_types} summarizes the differences between white-, gray-, and black-box scenarios, clearly displaying the amount of information the attacker can access in each of them. In this sense, the gray-box scenario assumes a compromise between what the attacker knows and the effect of the attacked images.

\IEEEpubidadjcol
This paper also presents an approach to duly defend against the proposed gray-box scenario, named Learning Image Similarity Adversa\textit{r}ial Defense (LISArD), which can also be applied to white-box settings without depending on adversarial examples. LISArD relates the similarity between clean and perturbed images by calculating the cross-correlation matrix between the embeddings of these images and using the loss to approximate this matrix to the identity while teaching the model to classify objects correctly. The goal of this approach is to reduce the effect of perturbations, motivating the model to recognize the clean and perturbed images as similar. This paper contributions can thus be summarized as follows:
\begin{itemize}
    \item It introduces the first gray-box testing framework, solely based on the architecture and data used to train a network, which is more realistic than white-box scenarios;
    \item It presents a defense mechanism that helps standard networks to be robust against gray-box attacks, without additional training epochs, parameters, and adversarially attacked images;
    \item Ablation studies and experimental evaluation demonstrate LISArD is the most robust against gray-box attacks, without increasing training cost, and has the least performance decrease in white-box scenarios.
\end{itemize}

The remaining of the paper is structured as follows: section~2 discusses the related works, namely \textit{Adversarial Distillation} and \textit{Adversarial Purification}; section~3 describes preliminary concepts, provides LISArD formal definition, and justifies the attack selection; section~4 reports the experimental setup, ablation studies and performance analysis, accompanied by a discussion; finally, Section~5 concludes the paper.

\section{Related Work}
\label{sec:related-work}

\noindent\textbf{White-box Adversarial Attacks}.
L-BFGS~\cite{szegedy2014intriguing} was the first proposed adversarial attack that demonstrated how simple perturbations could affect the DNNs performance.
Fast Gradient Sign Method (FGSM)~\cite{goodfellow2015explaining} is a one-step method that uses the model cost function, the gradient, and the radius epsilon to search for perturbations.
Jacobian-based Saliency Maps (JSM)~\cite{papernot2016limitations} explore the forward derivatives and construct the adversarial saliency maps.
Gradient Aligned Adversarial Subspace (GAAS)~\cite{tramer2017space} estimates the dimensionality of the adversarial subspace using the first-order approximation of the loss function.
Sparse and Imperceivable Adversarial Attacks (SIAA)~\cite{croce2019sparse} create sporadic and imperceptible perturbations by applying the standard deviation of each color channel in both axis directions.
DeepFool~\cite{moosavi2016deepfool} is an iterative attack that stops when the minimal vector orthogonal to the hyperplane representing the decision boundary is found.
SmoothFool (SF)~\cite{dabouei2020smoothfool} is an iterative algorithm that uses DeepFool to calculate the initial perturbation and smoothly rectifies the resulting perturbation until the adversarial example fools the classifier.
Projected Gradient Descent (PGD)~\cite{madry2018towards} is an iterative attack that uses saddle point formulation to find a strong perturbation.
Momentum Iterative FGSM (MI-FGSM)~\cite{dong2018boosting} introduces momentum into the Iterative FGSM (I-FGSM).
Auto-Attack~\cite{croce2020reliable} is a set of attacks to evaluate the networks, proposing the APGD-CE (i.e., PGD using Cross-Entropy (CE)), and APGD-DLR (i.e., PGD using Difference of Logits Ratio (DLR)) attacks. These techniques are are combined with Fast Adaptive Boundary (FAB)~\cite{croce2020minimally}, used to minimize the norm of the adversarial perturbations, and the Square Attack~\cite{andriushchenko2020square}, a query-efficient black-box attack. LISArD proposes using white-box attacks against models with the same architecture and data as the target, but without assuming the attacker can access this target model, making our approach more suitable to deal with realistic scenarios.

\textbf{Adversarial Distillation}.
Defensive Distillation (DD)~\cite{papernot2016distillation}, and its extension~\cite{papernot2017extending}, were the first methods to demonstrate the usefulness of distillation to defend against adversarial examples.
Robust Self-Training (RST)~\cite{carmon2019unlabeled} uses a standard supervised approach to obtain pseudo-labels and feed them into another network that targets adversarial robustness.
Adversarially Robust Distillation (ARD)~\cite{goldblum2020adversarially} performs distillation using an adversarially trained network as the teacher.
Introspective Adversarial Distillation (IAD)~\cite{zhu2021reliable} evaluates the robustness of the teacher network considering both the student and teacher labels.
Robust Soft Label Adversarial Distillation (RSLAD)~\cite{zi2021revisiting} uses robust soft labels produced by a teacher network to supervise the student training on natural and adversarial examples.
Low Temperature Distillation (LTD)~\cite{chen2021ltd} considers low temperature in the teacher network and generates soft labels that can be integrated into existing works.
Robustness Critical Fine-Tuning (RiFT)~\cite{zhu2023improving} introduces the module robust criticality metric to fine-tune the less robust modules to adversarial perturbations.
Adaptive Adversarial Distillation (AdaAD)~\cite{huang2023boosting} involves the teacher model in the optimization process by interacting
with the student model to search for the inner
results adaptively.
Information Bottleneck Distillation (IBD)~\cite{kuang2024improving} uses soft-label distillation to increase the mutual information between latent features and predictions and transfers relevant knowledge from the teacher
to the student to reduce the mutual information between the input and latent features.
Fair Adversarial Robustness Distillation (FairARD)~\cite{yue2024revisiting} ensures robust fairness of the student by increasing the weights for naturally more difficult classes.
PeerAiD~\cite{jung2024peeraid} trains a peer network on
the adversarial examples generated for the student network, simultaneously training the student and peer network.
Dynamic Guidance Adversarial Distillation (DGAD)~\cite{park2025dynamic} corrects teacher and student misclassification on clean and adversarially perturbed images. LISArD also does not include additional models during the inference phase, without involving Adversarial Training (AT) and larger previously trained models, thus being a more reliable approach for various domains.

\begin{figure*}[!t]
    \centering
    \includegraphics[width=0.9\linewidth]{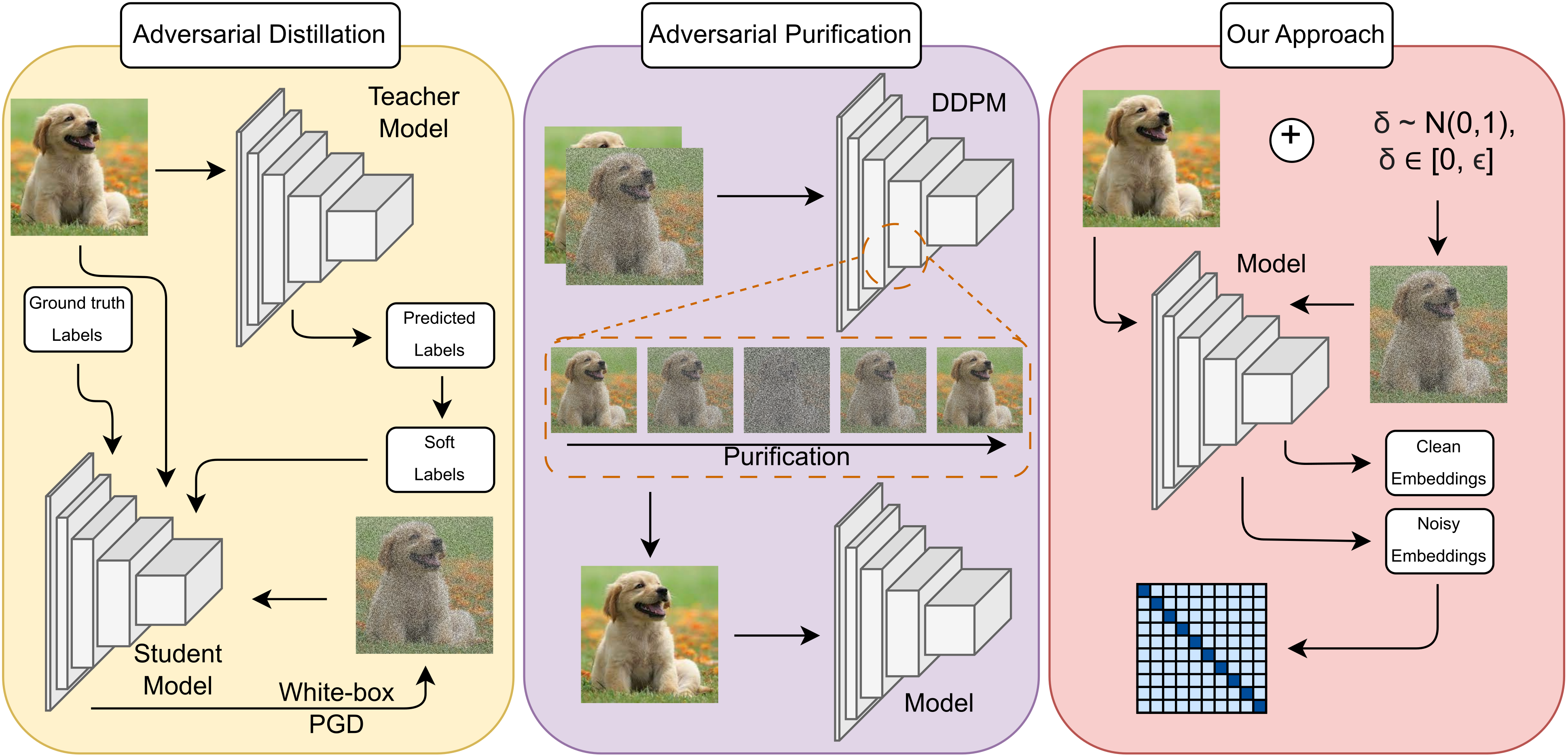}
    \caption{Types of approaches commonly used to defend against adversarial attacks. The Teacher Model refers to a previously trained model, usually bigger than the Student Model, that aids the latter by providing soft labels. The DDPM refers to a Denoising Diffusion Probabilistic Model (a generative model) that uses noise and denoise to produce a ``purified'' image.}
    \label{fig:approaches}
\end{figure*}

\textbf{Adversarial Purification}.
Yoon \textit{et al.}~\cite{yoon2021adversarial} propose using an Energy-Based Model with Denoising Score-Matching to purify perturbed images quickly.
For the first time, diffPure~\cite {nie2022diffusion} uses DDPM to remove the adversarial perturbations from the input images.
Guided Diffusion Model for Adversarial Purification (GDMAP)~\cite{wu2022guided} gradually denoises pure Gaussian noise with guidance to an adversarial image.
APuDAE~\cite{kalaria2022towards} uses Denoising AutoEncoders~\cite{vincent2008extracting} to purify the adversarial examples in an adaptive way, improving the accuracy of target networks.
DensePure~\cite{chen2022densepure} uses different random seeds to get multiple purified images, which are fed to the classifier, and its final prediction is based on majority voting.
Wang \textit{et al.}~\cite{wang2023better} uses better diffusion models~\cite{karras2022elucidating} to demonstrate that higher efficiency and quality diffusion models translate into better robust accuracy.
Lee \textit{et al.}~\cite{lee2023robust} propose a gradual noise-scheduling strategy that improves the robustness of diffusion-based purification.
Feature Purification Network (FePN)~\cite{cao2023fepn} is an adversarial learning mechanism that learns robust features by removing non-robust features from inputs while reconstructing high-quality clean images.
DifFilter~\cite{chen2024diffilter} uses a score-based method to improve the data distribution of the clean samples.
DiffAP~\cite{zhang2024random} uses conditional guidance to ensure prediction consistency between the purified and clean images. 
MimicDiffusion~\cite{song2024mimicdiffusion} approximates the purification process of adversarial examples and clean images by using Manhattan distance and two guidances. Adversarial Purification is the most efficient defense approach for DNNs, but it comes at the cost of high computational resources, while LISArD is able to protect different architectures in various setups without requiring additional training overhead.

\section{LISArD Methodology}
\label{sec:proposed_model}

\subsection{Adversarial Context and Preliminary}

\noindent\textbf{White-box Issues}. The white-box attacks are the strongest attacks for a specific model, yet if its training method slightly diverges, the same perturbations no longer have the identical effect as the model that was used to generate the adversarial samples. Furthermore, the white-box scenario requires that the attacker has access to the implementation/code of the model, which might not be realistic in most cases, since the attacker will rarely have access to the code of deployed models.

\textbf{Black-box Problems}. The black-box attacks are mainly focused on generating perturbations based on a low amount of knowledge, reducing the effect of the adversarial samples when compared to the white-box. However, the former attacks are more viable since the attacker only needs to know pairs of images and answers given by the target model to generate the perturbations. The black-box scenario can be considered as the most generic nowadays, since it does not require any information about the model, being potentially applicable to any available system exposing an DNN. Nevertheless, in this scenario, the attacker does not benefit from additional details of the target model, which hinders the probability of success.

\textbf{Proposed Solution}. We propose an alternative scenario in which the attacker knows the architecture of the model and dataset used to train it but does not have access to the gradients of the model. This information is usually accessible in papers or descriptive pages of the model, which can help the attacker to achieve stronger perturbations. This scenario is more realistic than white-box by compromising the amount of knowledge needed and the influence of the perturbations on the target model. LISArD considers using white-box attacks to generate adversarial samples against a model that uses the same architecture and dataset as the target model.

\textbf{Types of Approaches}. The two approaches described in the specialized literature and the approach proposed herein are summarized in Figure~\ref{fig:approaches}, which highlights the increased resources for performing \textit{Adversarial Distillation} and \textit{Purification} compared to LISArD. \textit{Adversarial Distillation} requires the usage of an additional previously trained model (Teacher) to aid in teaching the resilient network (Student), and \textit{Adversarial Purification} involves the training of a generative model (DDPM) to remove the adversarial noise during the inference phase (Purification). LISArD considers its attack scenario as a gray-box, meaning that the attacker only has partial knowledge about the target model. Thus, training to perceive noisy images (created by adding random Gaussian noise) similar to clean images can aid in defending against this type of attack.

\begin{figure}[!t]
    \centering
    \includegraphics[width=0.9\linewidth]{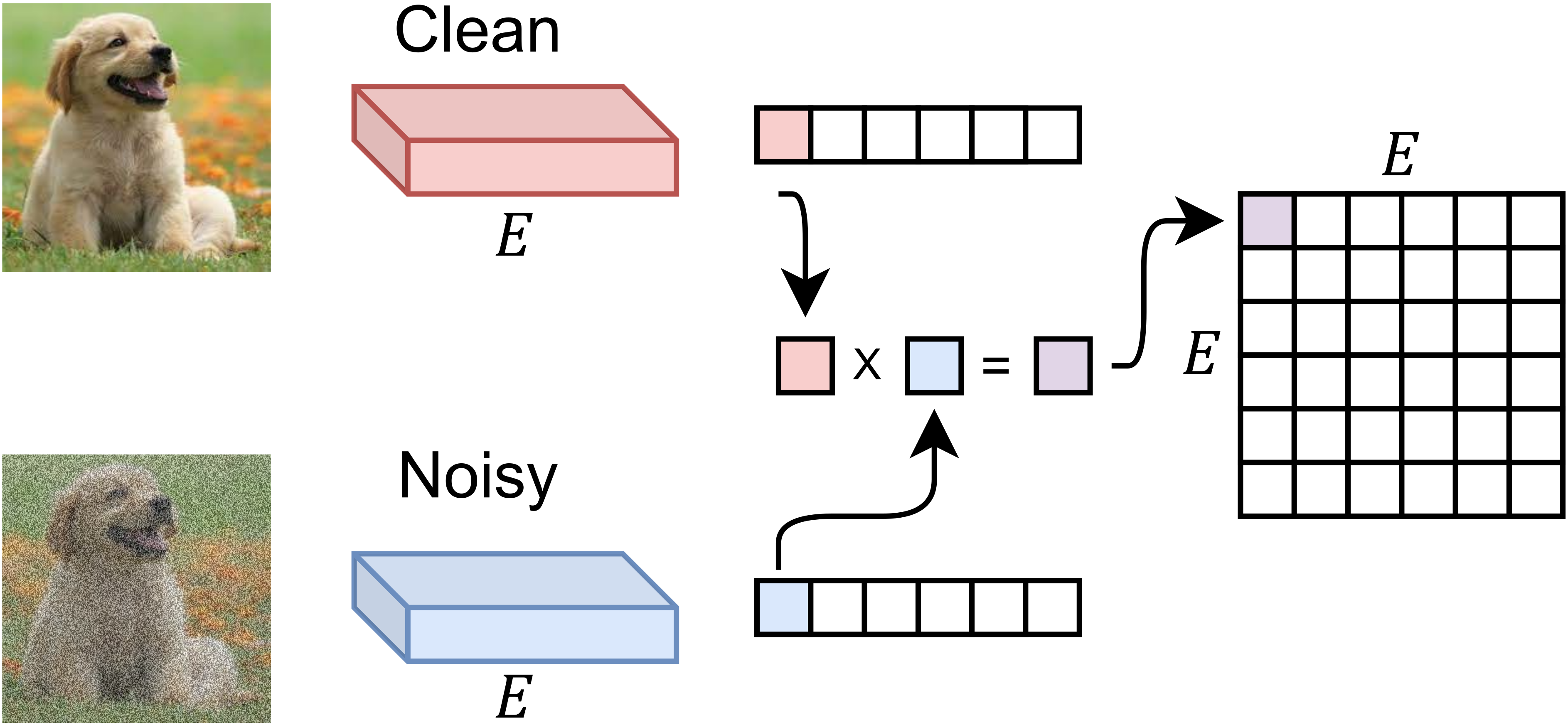}
    \caption{Overview of the conversion from embeddings to a matrix in the Learning Image Similarity component. $E$ refers to the size of the embeddings, which vary depending on the selected model.}
    \label{fig:learning_image_similarity}
\end{figure}

\subsection{Image Similarity and Importance}

\noindent\textbf{Motivation}. Learning Image Similarity (LIS) is based on the idea that an image containing a reduced amount of noise does not affect the object represented in that image. Barlow Twins~\cite{zbontar2021barlow} proposes a procedure to reduce the redundancy between a pair of identical networks in the context of self-supervised learning. LISArD utilizes the redundancy reduction approach to teach the model to identify the noisy and clean images as similar and improve robustness against gray-box and white-box attacks. 

\textbf{Embeddings to Matrix Conversion}. An overview of the LIS component, explaining the conversion process from embeddings to a matrix, which is used to achieve redundancy reduction between images is provided in Figure~\ref{fig:learning_image_similarity}. The embeddings with size $E$ are extracted before being fed to the classification layer, and each clean embedding is multiplied by each noisy embedding to obtain the cross-correlation matrix. Then, this cross-correlation matrix is approximated to the diagonal matrix to achieve a perfect correlation.

\begin{figure}[!t]
    \centering
    \includegraphics[width=0.9\linewidth]{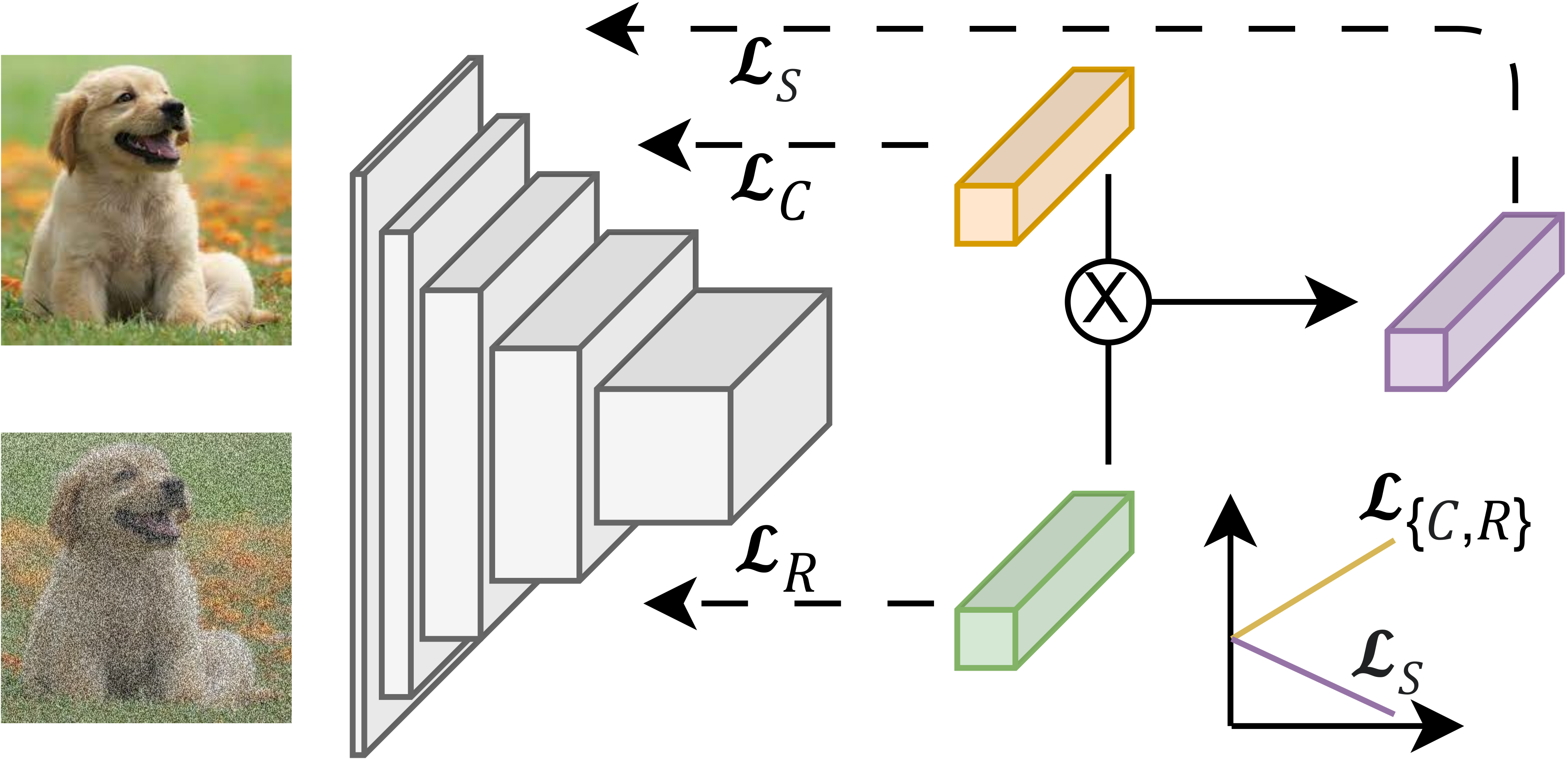}
    \caption{Overview of the LISArD~architecture. The clean and noisy images are fed to the model, and the inner product is calculated using their respective embeddings. Both clean (orange) and noisy embeddings (green) are used to predict each class using an adaptive weight loss between $\mathcal{L}_{C}$ and $\mathcal{L}_{R}$ and $\mathcal{L}_{S}$.}
    \label{fig:weighted_loss}
\end{figure}

\textbf{Weighted Training}. LISArD focuses on two main approaches: learning that two images are similar and simultaneously learning to classify the images, which motivates the usage of weighted training. Figure~\ref{fig:weighted_loss} explains how LISArD relates the LIS component with the classification one. As previously explained, the embeddings obtained from clean and noisy images are used in LIS while simultaneously being forwarded to the classification layer. The predictions are used in the (losses) $\mathcal{L}_{C}$ and $\mathcal{L}_{R}$ for the clean and noisy images, respectively, to train the classification component (the losses are better explained below). With this approach, we intend that the model initially concentrates on learning that two images represent the same object, but the final task is the classification of that object, justifying the initially increased importance of LIS and the gradually increasing importance of classification toward the end of training.

\subsection{Loss Function}

\noindent The LISArD consists of a new defense mechanism that does not cost significantly more than the standard training but provides the networks with robustness against gray-box and white-box adversarial attacks. It starts by generating random images for every batch, according to the following equation:

\begin{equation}
x_{R} = x_{C} + \sqrt{\mu} \cdot x_{N},
\end{equation}
where $x_{R}$ refers to the random image, $x_{C}$ refers to the clean image, and $\mu$ is the maximum amount of perturbation to be added to the image (simulating the $\epsilon$ from adversarial attacks). $x_{N}$ refers to the Gaussian noise with the same size as the clean image. Since we have two images that are given as input to the model, we have a classification loss for each of them. Formally, this loss is defined as the comparison between the predicted label and the ground truth via Cross-Entropy:

\begin{equation}
\mathcal{L}_{\{C,R\}} = (y~\text{log}(p) + (1 - y)~\text{log}(1 - p)),
\end{equation}
where $\mathcal{L}_{\{C,R\}}$ refers to either the clean image loss or random image loss, $p$ are the predicted labels for the images batch, and $y$ are the ground truth labels for the images batch. Another part of the loss function consists of the approximation between the embeddings of each input image. The following equation translates this process:

\begin{equation}
    \mathcal{L}_{S} = \sum_i (1 - M_{ii})^2 + \lambda \sum_i \sum_{j \neq i} M_{ij}^2,
\end{equation}
where $\lambda$ is a positive constant that balances the importance of the terms and $M$ is the cross-correlation matrix obtained by the embeddings of the two images along the batch:

\begin{equation}
    M_{ij} = \frac{ \sum_b z^A_{b,i} z^B_{b,j} }{ \sqrt{\sum_b (z^A_{b,i})^2} \sqrt{\sum_b (z^B_{b,j})^2}},
\end{equation}
where $b$ is the index for the batch samples and $i$, $j$ are the indexes for the elements of the matrix. $M$ is a square matrix with a size equal to the network output. Finally, the complete loss function is expressed by:

\begin{equation}
\begin{aligned}
\mathcal{L}_{} = \alpha~(\mathcal{L}_{C} + \mathcal{L}_{R}) + (1-\alpha)~\left(\frac{\mathcal{L}_{S}}{\tau}\right), 
\end{aligned}
\label{eq:final_loss}
\end{equation}
where $\mathcal{L}_{C}$, $\mathcal{L}_{R}$, and $\mathcal{L}_{S}$ refer to the losses for clean images, random images (defined in equation 2), and similarity approximation (defined in equation 3). $\tau$ refers to the temperature and $\alpha$ refers to the weight for classification, with $\alpha$ starting at 0.5 and incrementing to 1 throughout training, as follows:

\begin{equation}
\begin{aligned}
\alpha = \alpha_{0} + \delta(\varepsilon-1),
\end{aligned}
\end{equation}
where $\alpha_{0}$ is the starting coefficient, defined as 0.5, $\delta$ is the decay degree, set to $\frac{1}{400}$ and $\varepsilon$ refers to the training epoch.

\subsection{Selected Attacks and State-of-the-art}

\noindent FGSM~\cite{goodfellow2015explaining}, PGD~\cite{madry2018towards}, and AA~\cite{croce2020reliable} attacks are selected to evaluate LISArD and compare it with state-of-the-art. FGSM is a one-step adversarial attack that uses the gradients of the model, being a weaker white-box adversarial attack. PGD is a strong attack that many defenses still fail to overcome and has multiple iterations that increase its strength. AA consists of an ensemble of attacks containing white-box and black-box variants, allowing an evaluation in both settings, which increases the scope of our evaluation. AdaAD~\cite{huang2023boosting}, PeerAiD~\cite{jung2024peeraid}, and DGAD~\cite{park2025dynamic} are the approaches selected to compare with LISArD since these \textit{Adversarial Distillation} models achieve state-of-the-art performance in white-box settings and have available implementations.

\subsection{Implementation Details}

\noindent\textbf{Hardware.} The experiments were performed in a multi-GPU server containing seven NVIDIA A40 and an Intel Xeon Silver 4310 \symbol{`@} 2.10 GHz, with the Pop!\_OS 22.04 LTS operating system. The models were trained using a single NVIDIA A40 GPU without additional models running on the same GPU when presenting the total time or time per epoch results.

\textbf{Models}. In order to be as comprehensive as possible regarding the multiple proposal of architectures, we selected ResNet18~\cite{he2016deep}, ResNet50~\cite{he2016deep}, ResNet101~\cite{he2016deep}, WideResNet28-10~\cite{zagoruyko2016wide}, VGG19~\cite{simonyan2014very}, MobileNetv2~\cite{sandler2018mobilenetv2}, and EfficientNetB2~\cite{tan2019efficientnet} as our backbones. 
For all the datasets, the networks were trained using an SGD optimizer with a learning rate of 0.001, a momentum of 0.9, and a weight decay of 0.0005 during 200 epochs. 
We disregarded the training of Inceptionv3~\cite{szegedy2016rethinking} due to its need to increase the image size to 299x299, which would not be the same training and evaluation settings as other models.

\textbf{Ablation Studies}. Models were trained for 200 epochs using ResNet18 as the backbone architecture for all ablation studies and evaluated on the CIFAR-10 clean, FGSM, PGD, and AA datasets. The last three datasets were generated by applying the respective attack to a previously trained ResNet18 on CIFAR-10 clean.

\section{Experiments}

\subsection{Experimental Setup}

\noindent\textbf{Datasets}. 
The datasets are based on the most recent papers addressing adversarial defenses and their performance. 
The models are evaluated on CIFAR-10~\cite{krizhevsky2009learning} and CIFAR-100~\cite{krizhevsky2009learning}, consisting of 50 000 training and 10 000 testing images, and Tiny ImageNet~\cite{le2015tiny}, which has 100 000 training and 10 000 validation images, and is a subset of ImageNet comprising only 200 classes. These datasets are widely adopted in Adversarial Attacks in Object Recognition specialized literature.

\textbf{Gray-box Attacks.} Each selected architecture was trained in the CIFAR-10~\cite{krizhevsky2009learning}, CIFAR-100~\cite{krizhevsky2009learning}, and Tiny ImageNet~\cite{le2015tiny} datasets and typical models with the best clean accuracy were selected. 
The weights of these models are then used to generate the adversarial images. For all the attacks, the perturbation constraint was set to $\epsilon = 8/255$ for CIFAR-10 and CIFAR-100 and $\epsilon = 4/255$ for Tiny ImageNet. The considered attacks were FGSM~\cite{goodfellow2015explaining}, 10 steps PGD~\cite{madry2018towards} with a step size of $2/255$, and AA~\cite{croce2020reliable} using the $L_\infty$ norm and standard version.

\textbf{Evaluation}. We use accuracy on natural test samples, denominated Clean Accuracy, and accuracy on adversarial test samples, represented by the attack name, to measure the performance of the model. The attacked datasets used to train the proposed approach are different from the ones used to evaluate LISArD.

\begin{table}[!tb]
    \centering
    \caption{Comparison of different training methods on gray-box settings on CIFAR-10. $S$, $I$, and $L$ refer to ResNet trained from scratch, with ImageNet pretraining, and LISArD, respectively.}
    \begin{tabular}{c|c|c|c|c}
        \multirow{2}{*}{\textbf{Model}} & \multicolumn{4}{c}{\textbf{Gray-box Accuracy}} \\
         & Clean & FGSM & PGD & AA \\
         \hline \hline
         ResNet$_{S}$ & 87.88 & 53.53 & 43.34 & 46.56 \\
         ResNet$_{I}$ & \textbf{94.43} & 38.21 & 3.25 & 7.13 \\
         ResNet$_{L}$ & 87.22 & \textbf{83.14} & \textbf{83.54} & \textbf{84.19} \\
         \hline
    \end{tabular}
    \label{tab:graybox_scenario}
\end{table}
\begin{table}[!tb]
    \centering
    \caption{Performance of multiple architectures on gray-box settings when trained from Scratch ($S$) and using LISAD ($L$), on CIFAR-10.}
    \begin{tabular}{c|c|c|c|c}
        \multirow{2}{*}{\textbf{Model}} & \multicolumn{4}{c}{\textbf{Gray-box Accuracy}} \\
         & Clean & FGSM & PGD & AA \\
         \hline \hline
         ResNet50$_{S}$ & 96.65 & 30.49 & 0.43 & 1.96 \\
         ResNet50$_{L}$ & 88.07 & 84.78 & 84.95 & 85.56 \\
         \hline
         ResNet101$_{S}$ & 96.25 & 45.51 & 6.60 & 9.84 \\
         ResNet101$_{L}$ & 87.64 & 84.86 & 85.03 & 85.26 \\
         \hline
         MobileNetv2$_{S}$ & 85.07 & 17.04 & 0.73 & 5.60 \\
         MobileNetv2$_{L}$ & 85.23 & 81.29 & 82.14 & 83.22 \\
         \hline
         WideRN28-10$_{S}$ & 89.52 & 33.36 & 4.11 & 8.58 \\
         WideRN28-10$_{L}$ & 88.43 & 80.03 & 80.81 & 83.15 \\
         \hline
         VGG19$_{S}$ & 91.61 & 15.01 & 0.08 & 2.35 \\
         VGG19$_{L}$ & 85.87 & 79.50 & 81.27 & 82.29 \\
         \hline
         EfficientNetB2$_{S}$ & 84.99 & 22.27 & 5.49 & 11.40 \\
         EfficientNetB2$_{L}$ & 77.67 & 72.01 & 73.53 & 74.26 \\
         \hline
    \end{tabular}
    \label{tab:diff_networks}
\end{table}

\subsection{Gray-box Settings}

\noindent\textbf{Gray-box Adversarial Attack Impact.} We start by studying if the gray-box scenario is also an issue for typical models. We report the gray-box accuracy for ResNet18 architecture using different training methods in Table~\ref{tab:graybox_scenario}. This scenario uses a ResNet18 model trained on the CIFAR-10 dataset to generate the adversarial images using the different attacks. Then, these images are given to other ResNet18 models to evaluate their robustness. Both models with and without pretrain are vulnerable to gray-box attacks, raising awareness for this more realistic type of attack. LISArD significantly helps to diminish the effect of gray-box attacks, which highlights the importance of image similarity for robust model defense.

\begin{figure}[!t]
    \centering
    \includegraphics[width=0.48\textwidth]{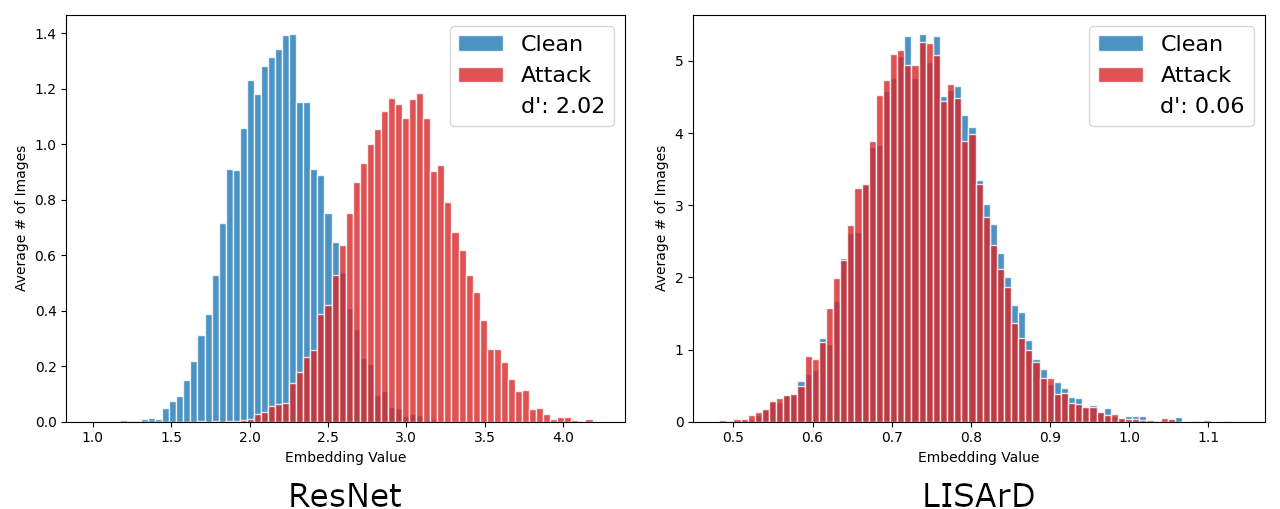}
    \caption{Comparison of the distributions for clean (blue) and attacked (red) images when considering a ResNet (left) and LISArD (right) for CIFAR-10. $\textit{d}'$ refers to the decidability measure, where values closer to 0 mean greater overlap between distributions.}
    \label{fig:Comparison_distribution}
\end{figure}

\textbf{Vulnerability of Different Architectures.} Since we are proposing a new training mechanism to diminish the effect of gray-box attacks, we need to evaluate if it can be applied to different networks. The gray-box accuracy for different architectures is presented in Table~\ref{tab:diff_networks}, ranging from 3.5M (MobileNetv2) to 143.7M (VGG19) parameters. The proposed method effectively helps to protect against gray-box attacks for multiple models with different architectures and number of parameters. Figure~\ref{fig:Comparison_distribution} highlights the LISArD approach of clean and perturbed images representing the same concept, translated by the overlap of the distributions of clean and attacked image embeddings. To objectively quantify this overlap of information, we use the decidability measure~\cite{daugman2000biometric} ($\textit{d}'$), which shows that LISArD effectively approximates the distributions of clean and attacked images ($\textit{d}'$ close to 0), increasing protection against the gray-box attacks without additional training effort.

\begin{figure}[!t]
    \centering
    \includegraphics[width=0.85\linewidth]{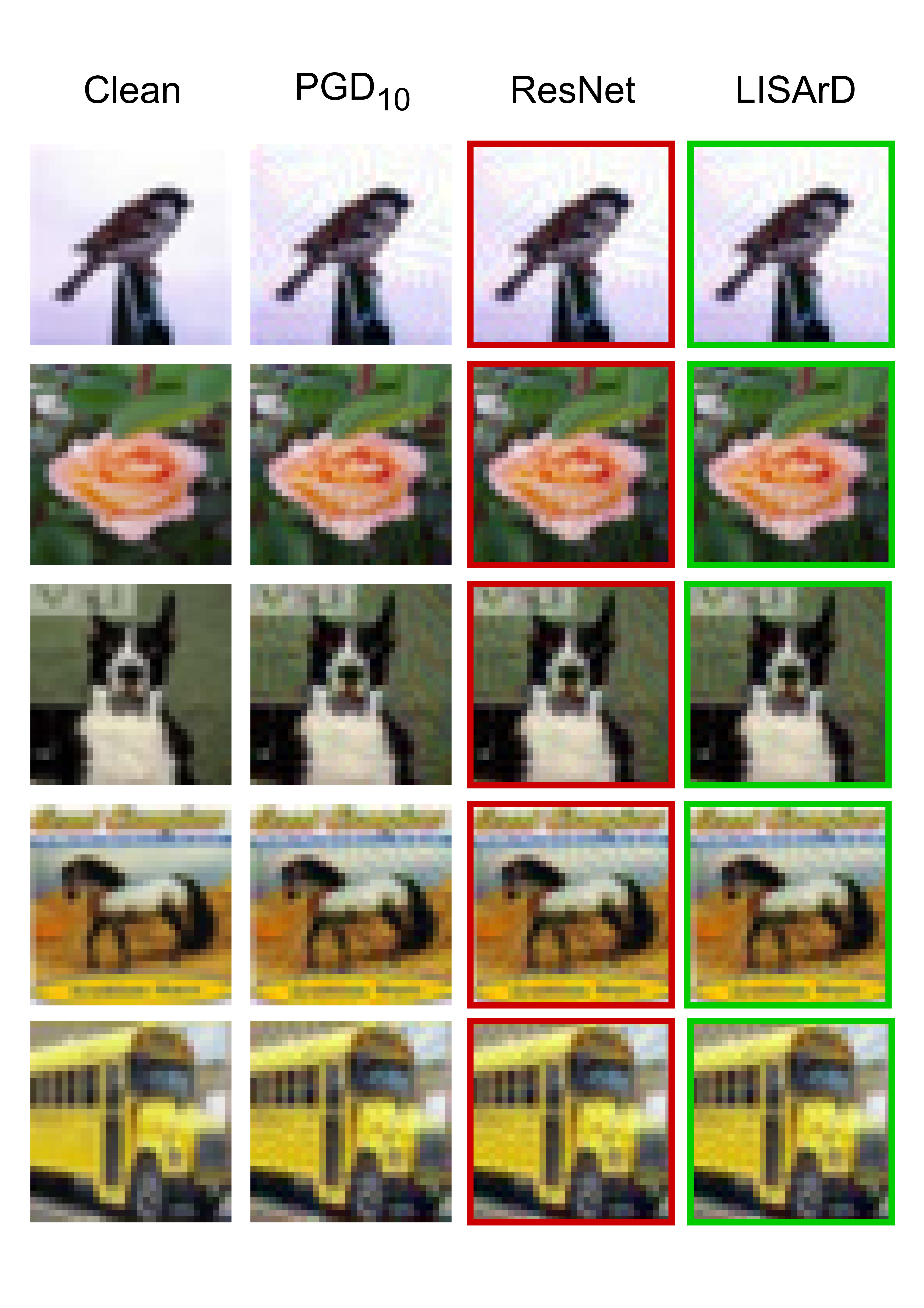}
    \caption{Clean and PGD$_{10}$ images and the effect of adversarial attacks on ResNet and LISArD trained networks for CIFAR-10 and CIFAR-100 datasets. Red and Green refer to incorrect and correct classifications, respectively.}
    \label{fig:comparison_CIFAR_Normal_LISAD}
\end{figure}

\begin{table*}[!tb]
    \centering
    \caption{Performance comparison with state-of-the-art approaches, with and without the inclusion of Adversarial Training (AT), on gray-box and white-box settings on CIFAR-10 and CIFAR-100.}
    \begin{tabular}{c|c|c|c|c|c|c|c|c|c|c}
        \multirow{2}{*}{\textbf{Dataset}} & \multirow{2}{*}{\textbf{Model}} & \multicolumn{4}{c}{\textbf{Gray-box Accuracy}} & \multicolumn{4}{c}{\textbf{White-box Accuracy}} & t/ep (min)\\
        & 
        & Clean & FGSM & PGD & AA
        & Clean & FGSM & PGD & AA \\
         \hline \hline
        \multirow{12}{*}{\textbf{CIFAR-10}}
         & AdaAD~\cite{huang2023boosting} 
         & 80.32 & 77.53 & 77.92 & 78.14
         & 85.58 & 60.85 & 56.40 & 51.37 & 09:52 \\
         & AdaAD wo/ AT 
         & 88.89 & 70.71 & 63.83 & 67.06
         & 88.89 & 37.93 & 1.39 & 0.11 & 09:46 \\
         & $\Delta$ 
         & \textbf{+8.57} & \underline{-6.82} & \underline{-14.09} & \underline{-11.08}
         & \underline{+3.31} & \textbf{-22.92} & -55.01 & -51.26 & -  \\
         \cline{2-11}
         & PeerAiD~\cite{jung2024peeraid} 
         & 84.38 & 81.88 & 82.30 & 82.63
         & 85.01 & 61.28 & 54.36 & 52.57 & 02:13 \\
         & PeerAiD wo/ AT 
         & 10.00 & 10.00 & 10.00 & 10.00
         & 10.00 & 10.00 & 10.00 & 10.00 & 02:07 \\
         & $\Delta$ 
         & -74.38 & -71.88 & -72.30 & -72.63
         & -75.01 & -51.28 & \underline{-44.36} & \underline{-42.57} & -  \\
         \cline{2-11}
         & DGAD~\cite{park2025dynamic} 
         & 87.50 & 84.91 & 85.47 & 85.83
         & 85.75 & 62.28 & 58.05 & 52.34 & 09:54 \\
         & DGAD wo/ AT 
         & 41.19 & 36.97 & 36.69 & 37.10
         & 37.32 & 3.77 & 0.08 & 0.00 & 09:48 \\
         & $\Delta$ 
         & -46.33 & -47.94 & -48.78 & -48.73
         & -48.43 & -58.51 & -57.97 & -52.34 & - \\
         \cline{2-11}
         & LISAD 
         & 80.42 & 78.19 & 78.48 & 78.54
         & 80.42 & 54.43 & 50.12 & 46.11 & 01:37 \\ 
         & LISAD wo/ AT 
         & 87.22 & 83.14 & 83.54 & 84.19
         & 87.22 & 27.47 & 13.92 & 11.84 & 00:25 \\
         & $\Delta$ 
         & \underline{+6.80} & \textbf{+4.95} & \textbf{+5.06} & \textbf{+5.65}
         & \textbf{+6.80} & \underline{-26.96} & \textbf{-36.20} & \textbf{-34.27} & - \\
        \hline \hline
        \multirow{12}{*}{\textbf{CIFAR-100}}
         & AdaAD~\cite{huang2023boosting} 
         & 61.82 & 58.91 & 58.83 & 59.55
         & 62.19 & 35.33 & 32.52 & 26.74 & 09:53 \\
         & AdaAD wo/ AT 
         & 67.85 & 51.39 & 52.54 & 54.65 
         & 67.85 & 23.20 & 3.47 & 1.07 & 09:47 \\
         & $\Delta$ 
         & \underline{+6.03} & \underline{-7.52} & \underline{-6.29} & \underline{-4.90}
         & \underline{+5.66} & \textbf{-12.13} & -29.05 & -25.67 & - \\
         \cline{2-11}
         & PeerAiD~\cite{jung2024peeraid} 
         & 59.37 & 57.15 & 56.84 & 57.80
         & 59.35 & 34.41 & 29.69 & 27.33 & 02:10 \\
         & PeerAiD wo/ AT 
         & 1.00 & 1.00 & 1.00 & 1.00 
         & 1.00 & 1.00 & 1.00 & 1.00 & 01:59 \\
         & $\Delta$ 
         & -58.37 & -56.15 & -55.84 & -56.80 
         & -58.35 & -33.41 & \underline{-28.69} & -26.33 & - \\
         \cline{2-11}
         & DGAD~\cite{park2025dynamic} 
         & 63.26 & 60.77 & 60.07 & 61.33 
         & 63.24 & 36.09 & 33.68 & 27.66 & 09:55 \\
         & DGAD wo/ AT 
         & 25.71 & 22.45 & 22.35 & 22.71
         & 10.79 & 4.46 & 4.20 & 2.20 & 09:49 \\
         & $\Delta$ 
         & -37.55 & -38.32 & -37.72 & -38.62
         & -52.45 & -31.63 & -29.48 & \underline{-25.46} & - \\
         \cline{2-11}
         & LISAD 
         & 54.30 & 52.20 & 52.29 & 52.52
         & 54.30 & 27.84 & 25.33 & 21.13 & 01:38 \\
         & LISAD wo/ AT 
         & 59.47 & 56.00 & 55.72 & 56.91 
         & 59.47 & 12.12 & 6.70 & 5.71 & 00:17 \\
         & $\Delta$ 
         & \textbf{+18.01} & \textbf{+3.80} & \textbf{+3.43} & \textbf{+4.39} 
         & \textbf{+18.01} & \underline{-15.72} & \textbf{-18.63} & \textbf{-15.42} & - \\
         \hline
    \end{tabular}
    \label{tab:SOTA_comparison_CIFAR}
\end{table*}

\textbf{Gray-box Adversarial Examples}. To further understand the impact of gray-box settings, we demonstrate scenarios where a typical network fails to correctly classify the object, in Figure~\ref{fig:comparison_CIFAR_Normal_LISAD}. A typical network has difficulty in rightly classifying images with clearly outlined objects or with almost no background, only by adding perturbations that do not impair Human decision (third column in the figure). This confirms the hypothesis that typical networks are also vulnerable to gray-box attacks, which is mitigated when using LISArD, where the networks can now correctly classify the same images (fourth column in the figure), highlighting the importance of image similarity-based training to provide increased robustness.

\subsection{Adversarial Robustness}

\noindent To the best of the authors' knowledge, no previous works in the literature propose evaluating the networks in the gray-box scenario. Therefore, images solely for the purpose of evaluation were generated to ensure a fair comparison between the different approaches, effectively assuring that the models had not previously seen these images. The adversarial robustness is evaluated on CIFAR-10 and CIFAR-100 in Table~\ref{tab:SOTA_comparison_CIFAR} and on Tiny ImageNet in Table~\ref{tab:sota_comparison_TinyImageNet}.

\textbf{AT Effect on Model Performance}. State-of-the-art models typically include adversarial samples in training to increase model robustness, which gives an inherent advantage in white-box settings. To assess the resilience in both scenarios (gray-box and white-box) and to make a fair comparison with LISArD, we consider the settings with and without AT for the different models in our experiments. Table~\ref{tab:SOTA_comparison_CIFAR} compares LISArD and \textit{Adversarial Distillation} approaches, with and without AT during the training phase, showing that the models are highly dependent on AT examples to perform in white-box settings and are not as resilient in gray-box settings without these samples. The results also show that to improve resilience against white-box attacks, the Adversarial Distillation approaches are highly reliant on including AT and not on proposing a different type of approach.

\textbf{Gray-box \textit{vs.} White-box}. 
The difference in Clean accuracy between the gray-box and white-box evaluation for AdaAD, PeerAiD, and DGAD is: 1) the former is obtained from models trained according to the author's available implementations, and 2) the latter were obtained directly from the paper reported experiments~\cite{huang2023boosting,jung2024peeraid,park2025dynamic}.
In Table~\ref{tab:SOTA_comparison_CIFAR}, when comparing the results referring to gray-box attacks (4th, 5th, and 6th columns) with the ones from white-box attacks (8th, 9th, and 10th columns), we can note that the pattern regarding the strength of the attack is the same. In both scenarios, AA is the strongest, followed by PGD, and FGSM, respectively, suggesting that the effectiveness in the white-box scenario is also transferred to the gray-box scenario. This shows that the selected settings are representative of strong attacks, and the gray-box scenario has a difficulty aligned with the white-box one.

\textbf{State-of-the-art Methods}.
When evaluating the gray-box scenario, AdaAD is the second most resilient defense when removing the AT due to their reduced reliance on the labels from adversarial samples. Additionally, the same model is robust against FGSM in the white-box scenario, suggesting that learning from a teacher might be reliable for single-step attacks.
The remaining Adversarial Distillation methods rely heavily on including adversarial samples during the training phase to provide robustness against both gray-box and white-box attack scenarios.
PeerAiD without including AT performs similarly to a model with random predictions, which suggests that this defense is highly (or completely) dependent on the inclusion of AT to train a resilient model, justified by the removal of the ground-truth labels during the training phase. The results show that all analyzed models underperform greatly when removing AT and/or moving to gray-box settings, highlighting the limitations of existing approaches and their lack of robustness to perform in various conditions (aside from white-box settings).

\textbf{Overall LISArD Perfomance}.
LISArD is the least time-consuming method whilst offering the best overall resilience against attacks in a gray-box scenario due to its similarity learning relying solely on mathematical operations without including additional models.
For both datasets, LISArD shows a decrease in accuracy for all the attacks when including the AT approach in the gray-box scenario, suggesting that including adversarial samples in the training phase weakens the generalization capability.
Nevertheless, the inclusion of AT diminishes the clean accuracy of all models, which does not happen with the same impact when using the proposed defense.
This shows that LISArD performs the best in gray-box settings and is the most resilient defense in white-box settings when the adversarial samples are removed from the training stage, as shown by the $\Delta$ for both gray-box and white-box.

\begin{table}[!tb]
    \centering
    \caption{Comparison with state-of-the-art approaches on gray-box settings, using ResNet18 architecture on Tiny ImageNet.}
    \begin{tabular}{c|c|c|c|c}
        \multirow{2}{*}{\textbf{Model}} & \multicolumn{4}{c}{\textbf{Gray-box Accuracy}} \\
        & Clean & FGSM & PGD & AA \\
        \hline \hline
         Standard & 
         67.84 & 11.45 & 3.39 & 11.07 \\
         LISAD$_{R}$ & 
         54.64 & 50.50 & 48.52 & 51.73 \\
        \hline
    \end{tabular}
    \label{tab:sota_comparison_TinyImageNet}
\end{table}

\textbf{Tiny ImageNet}. To demonstrate the applicability of LISArD to larger and diversified datasets, we display the results for Tiny ImageNet in Table~\ref{tab:sota_comparison_TinyImageNet}. It was not possible to provide the results for AdaAD, PeerAiD, and DGAD, because the available implementations did not provide enough details on how to train for Tiny ImageNet. Nonetheless, we compare the proposed approach with a standardly trained network, showing that the former has a greater capacity to resist gray-box attacks despite the increase of data and labels.

\begin{table}[!tb]
    \centering
    \caption{Comparison of the effect of using different mechanisms to generate images on gray-box settings, using ResNet18 architecture, on CIFAR-10.}
    \begin{tabular}{c|c|c|c|c|c}
        \multirow{2}{*}{\textbf{Model}} & \multicolumn{4}{c}{\textbf{Gray-box Accuracy}} & \multirow{2}{*}{\textbf{Time (h)}} \\
         & Clean & FGSM & PGD & AA & \\
         \hline \hline
         FGSM & 64.34 & 31.22 & 12.80 & 15.84 & 2:05:50 \\ 
         PGD & 74.26 & 32.08 & 17.58 & 39.54 & 2:11:23 \\
         AA & 68.83 & 41.89 & 36.21 & 38.79 & 2:21:26 \\
         Random & \textbf{87.22} & \textbf{83.14} & \textbf{83.54} & \textbf{84.19} & \textbf{1:26:43} \\
         \hline
    \end{tabular}
    \label{tab:approximation_using_diff_imgs}
\end{table}
\begin{table}[!tb]
    \centering
    \caption{Ablation study regarding the considered losses and optimizer, using ResNet18 architecture, on CIFAR-10. Optim refers to the used optimizer and $\mathcal{L}_{C}$ and $\mathcal{L}_{R}$ refer to the clean and random images classification losses, respectively.}
    \begin{tabular}{c|c|c|c|c|c|c|c}
    \multirow{2}{*}{Optim} & $\mathcal{L}_{C}$ & $\mathcal{L}_{R}$ & \multicolumn{4}{c}{\textbf{Gray-box Accuracy}} \\
     & & & Clean & FGSM & PGD & AA \\
    \hline \hline
          \multirow{3}{*}{LARS}
          & $\checkmark$ & $\times$
          & 57.89 & 56.57 & 53.41 & 53.72  \\
          & $\times$ & $\checkmark$
          & 71.90 & 69.94 & 69.26 & 66.19 \\
          & $\checkmark$ & $\checkmark$
          & 72.64 & 69.68 & 68.35 & 69.01  \\
          \hline
          \multirow{3}{*}{SGD}
          & $\checkmark$ & $\times$
          & 85.96 & 65.69 & 61.96 & 65.62  \\
          & $\times$ & $\checkmark$
          & 84.00 & 81.33 & 81.42 & 81.96 \\
          & $\checkmark$ & $\checkmark$
          & \textbf{87.22} & \textbf{83.14} & \textbf{83.54} & \textbf{84.19} \\
         \hline
    \end{tabular}
    \label{tab:loss_ablation_studies}
\end{table}

\subsection{Ablation Studies}

\begin{figure}[!t]
    \centering
    \includegraphics[width=0.95\linewidth]{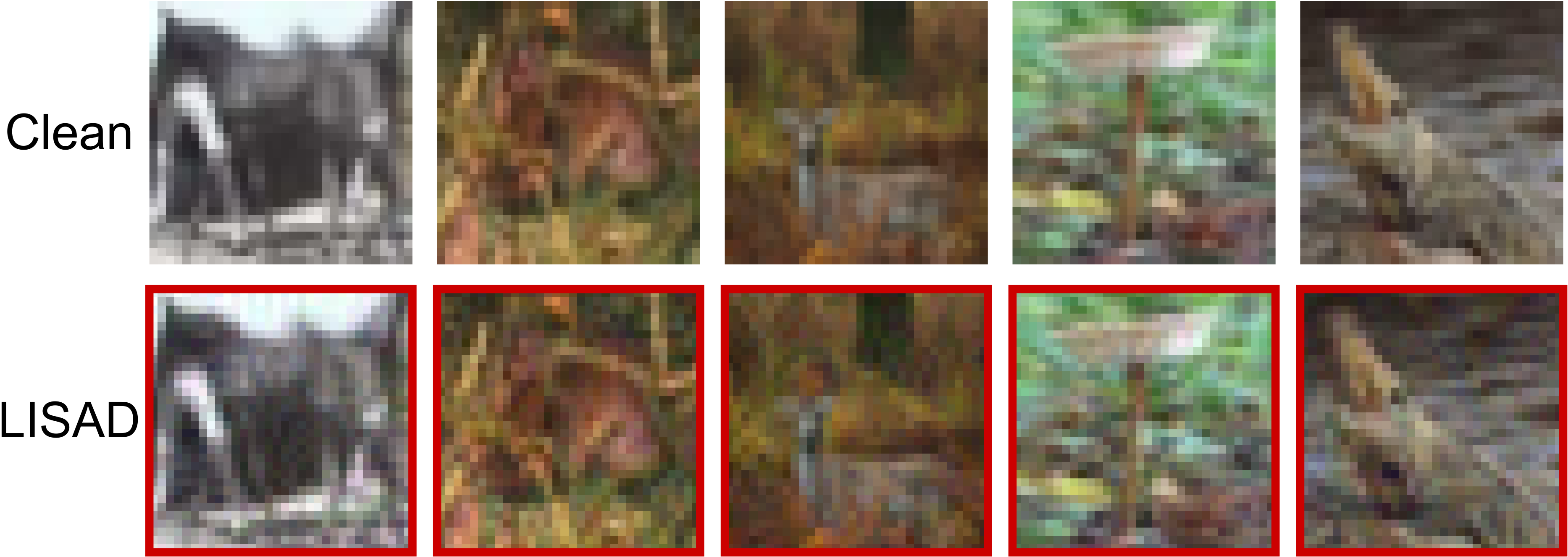}
    \caption{LISArD misclassification for CIFAR-10 and CIFAR-100 datasets, showing the difficulty to correctly classify objects that are blended with the background. Red refers to incorrect classifications.}
    \label{fig:Failures_CIFAR_LISAD}
\end{figure}

\noindent The loss function displayed in equation~\ref{eq:final_loss} was altered in multiple ways to find the adequate method for both learning image similarity and classification, considering Random as the image generation mechanism. The results for the ablation studies are displayed in Tables~\ref{tab:approximation_using_diff_imgs},~\ref{tab:loss_ablation_studies}, and~\ref{tab:components_ablation_studies}, and Figure~\ref{fig:Failures_CIFAR_LISAD} illustrates some scenarios where LISArD is unable to correctly classify the object.

\textbf{Different Image Generation Mechanisms}. Since LISArD intends to train a model to learn to approximate the noisy images to the clean images, the first ablation consists of evaluating the mechanism used to generate the noise images. Table~\ref{tab:approximation_using_diff_imgs} indicates the results for these evaluations, considering the FGSM, PGD, and AA attacks and adding Gaussian Noise (Random), with the loss function according to the one in equation~\ref{eq:final_loss}. As can be observed, the FGSM image generation fails to provide robustness against PGD and AA, suggesting the former is unsuccessful against multiple-step attacks. Although the PGD image generation improves the resilience against AA and PGD, it still performs less than AA, with the latter not significantly increasing the time cost. The increased performance observed in AA might be related to including multiple-step and black-box attacks in the image generation process, increasing the model generalization capability. Finally, the results show that adding white noise (Random) grants the best generalizing capability to the models for all the considered attacks whilst being the least resource-consuming because the images are generated without accessing the model gradients.

\textbf{Loss Function and Optimizer}. We start by evaluating the adequate optimizer for the main objectives of LISArD and if all the terms in the previously mentioned equation are necessary, as shown in Table~\ref{tab:loss_ablation_studies}. Since LISArD uses a training batch of 2048, we first explored the Layer-wise Adaptive Rate Scaling (LARS), which is an optimizer commonly used in greater-dimension batch sizes. However, the results demonstrate that using LARS is not the most effective means to make the classification component learn, leading to a performance significantly lower than a standard-trained network in clean accuracy (4$^{th}$ row in Table~\ref{tab:loss_ablation_studies}). Therefore, we opted to use the Stochastic Gradient Descent (SGD) as an optimizer, which is typically used in the literature to train the models (specifically for object recognition) and demonstrated overall better results than LARS. When considering only classifying the noisy images, the model performs better in the attack scenario but decreases performance for the clean images. On the other hand, solely classifying clean images demonstrates better results in clean accuracy. Thus, we opted for a conjunction between clean and noisy image classification, which exhibits the best results in overall accuracy.

\textbf{Loss Component Variation}. LISArD considers gradual learning throughout the 200 epochs, with greater initial importance given to the image similarity part and gradually increasing the importance of classification through the inclusion of $\alpha$. Additionally, temperature ($\tau$) was included in LISArD due to its proven increase in classification accuracy, specifically for \textit{adversarial distillation}~\cite{chen2021ltd,huang2023boosting}. Table~\ref{tab:components_ablation_studies} displays the results for balancing the classification and similarity components and including a temperature element. It is possible to observe that $\alpha$ significantly impacts the resilience of the model against adversarial attacks, reinforcing the significance of the image similarity component in that matter. The temperature $\tau$ is relevant to improve the results in both clean and attacked scenarios, as previously shown in the literature.

\begin{table}[!tb]
    \centering
    \caption{Ablation study regarding the considered components, using ResNet18 architecture, on CIFAR-10.}
    \begin{tabular}{c|c|c|c|c|c}
    \multirow{2}{*}{Component} & \multicolumn{4}{c}{\textbf{Gray-box Accuracy}} \\
    & Clean & FGSM & PGD & AA \\
    \hline \hline
    wo/ $\alpha$ and wo/ $\tau$ 
    & 68.61 & 67.13 & 66.48 & 66.73  \\
    w/ $\alpha$      
    & 75.01 & 72.89 & 72.26 & 72.80 \\
    w/ $\tau$      
    & 84.14 & 80.87 & 80.87 & 81.40 \\
    w/ $\alpha$ and w/ $\tau$      
    & \textbf{87.22} & \textbf{83.14} & \textbf{83.54} & \textbf{84.19} \\
    \hline
    \end{tabular}
    \label{tab:components_ablation_studies}
\end{table}

\textbf{Approach Limitations}. We display examples of scenarios that are challenging for LISArD in Figure~\ref{fig:Failures_CIFAR_LISAD}. LISArD fails to resist adversarial attacks when the pictures have the object masked in the image background. These scenarios are already hard to classify in a clean context, and their difficulty is exacerbated by adding perturbations to these images. As such, the seen underperformance of LISArD relates to the resemblance between the background and the object, making the classification inherently challenging, even when classifying the clean images.

\section{Conclusion}
\label{sec:conclusion}

\noindent This paper describes an evaluation framework based on the \emph{gray-box setting} that is more realistic than the typically used \emph{white-box} scenario, where the existing models do not perform reliably. We propose an adversarial defense mechanism for this setting (LISArD), which is simultaneously robust against white-box attacks, and does not depend on the inclusion of adversarial samples. This mechanism uses image similarity to instruct the model to recognize that images pairs regard the same object, while simultaneously inferring class information. The experiments show the vulnerability of pre-trained and scratch-trained networks to gray-box adversarial samples and point to the effectiveness of LISArD in increasing resilience against this type of samples. Also, state-of-the-art \textit{Adversarial Distillation} models cannot perform in white-box settings without the inclusion of AT. In the future, the injection of other types of noise (\textit{e.g.}, using fractional Gaussian noise with persistence, instead of white noise) will be subject of our further analysis. Finally, for realism purposes, we suggest evaluating the models in scenarios in which the attacker only knows the training data and the type of architecture.


\begin{IEEEbiography}[{\includegraphics[width=1in,height=1.25in,clip,keepaspectratio]{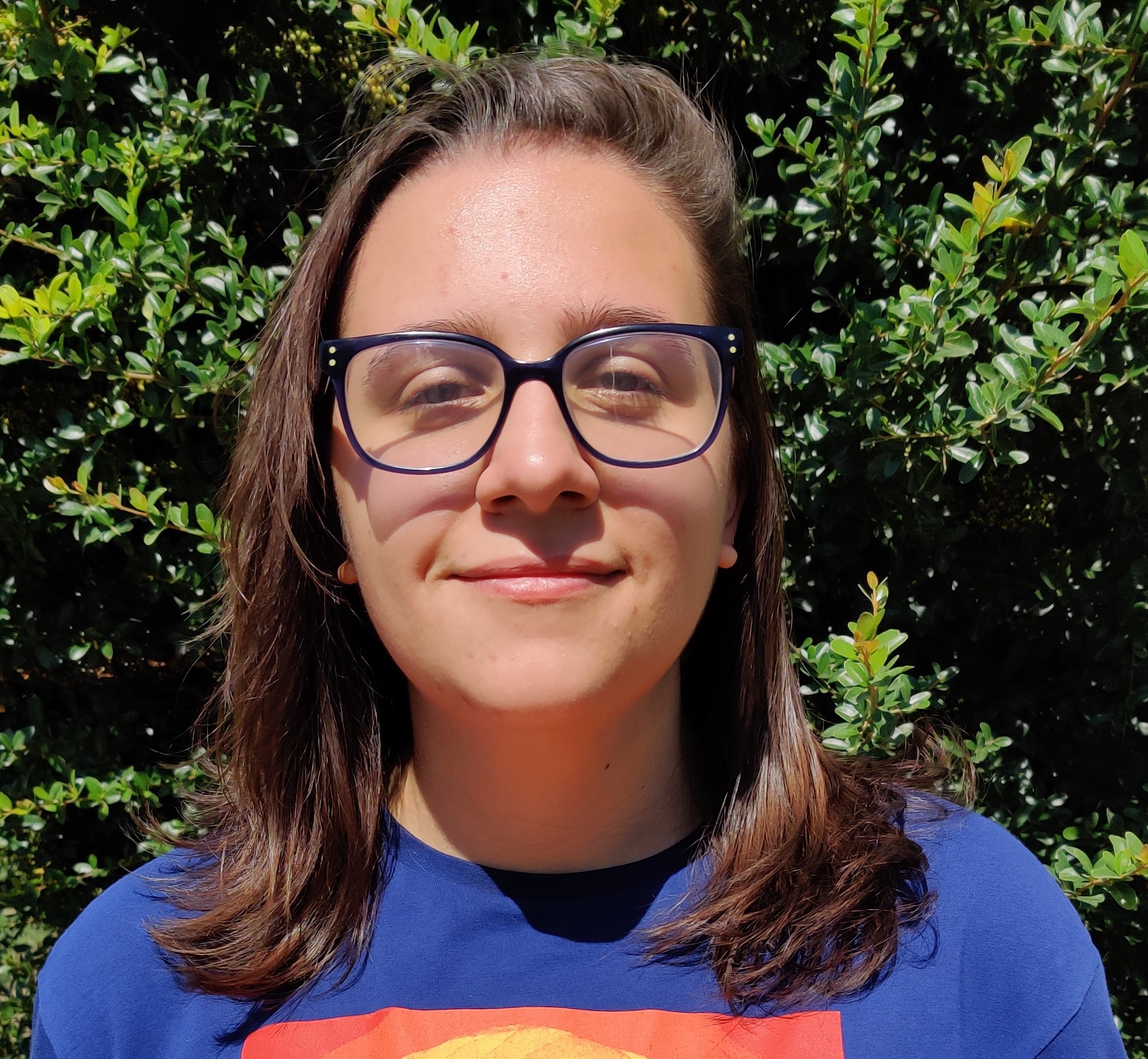}}]{Joana C. Costa} obtained her bachelor's and master's degree in Computer Science and Engineering from Universidade da Beira Interior (UBI) in 2019 and 2021, respectively. She is currently pursuing a Ph.D. degree, with an FCT (\textit{Funda\c{c}\~{a}o para a Ciência e a Tecnologia}) scholarship, in the field of Computer Vision and Adversarial Attacks.
\end{IEEEbiography}

\begin{IEEEbiography}[{\includegraphics[width=1in,height=1.25in,clip,keepaspectratio]{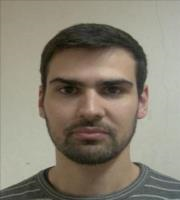}}]{Tiago Roxo} obtained a bachelor's degree in Computer Science and Engineering from Universidade da Beira Interior (UBI) in 2019 and is currently pursuing a Ph.D. degree, with an FCT (\textit{Funda\c{c}\~{a}o para a Ciência e a Tecnologia}) scholarship, in the field of Computer Vision and Artificial Intelligence.
\end{IEEEbiography}

\begin{IEEEbiography}[{\includegraphics[width=1in,height=1.25in,clip,keepaspectratio]{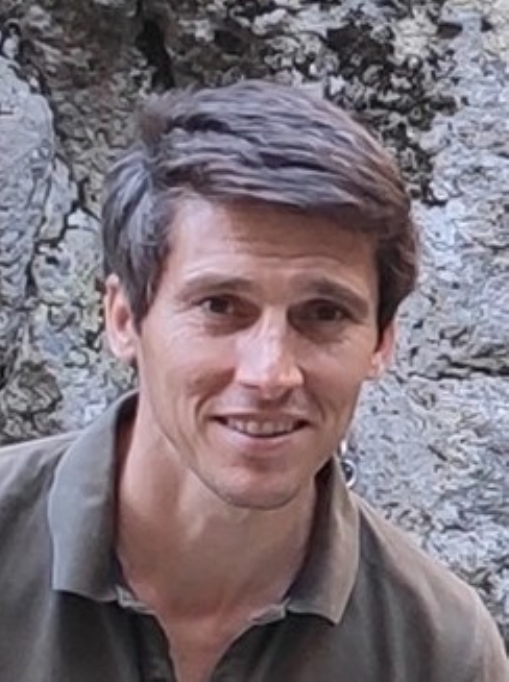}}]{Hugo Proen\c{c}a} (SM'12), B.Sc. (2001), M.Sc. (2004) and Ph.D. (2007) is a Full Professor in the Department of Computer Science, University of Beira Interior and has been researching mainly about biometrics and visual-surveillance. He was the coordinating editor of the IEEE Biometrics Council Newsletter and the area editor (ocular biometrics) of the IEEE Biometrics Compendium Journal. He is a member of the Editorial Boards of the Image and Vision Computing, IEEE Access and International Journal of Biometrics. Also, he served as Guest Editor of special issues of the Pattern Recognition Letters, Image and Vision Computing and Signal, Image and Video Processing journals. 
\end{IEEEbiography}

\begin{IEEEbiography}[{\includegraphics[width=1in,height=1.25in,clip,keepaspectratio]{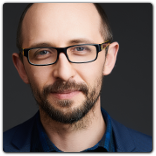}}]{Pedro R. M. In\'{a}cio}  (SM'15), B.Sc. in Mathematics/Computer Science (2005), and Ph.D. in Computer Science and Engineering (2009) is an associate professor of the Department of Computer Science at the University of Beira Interior (UBI), which he joined in 2010 and where he lectures subjects related with information assurance and (cyber)security. The Ph.D. work was performed in the enterprise environment of Nokia Siemens Networks Portugal S.A.

He is an IEEE senior member, an ACM professional member and a researcher of the Instituto de Telecomunicações (IT). His main research topics are information assurance and security, computer based simulation, and network traffic monitoring, analysis and classification. He frequently reviews papers for IEEE, Springer, Wiley and Elsevier journals. He is a member of the Technical Program Committees of flagship national and international workshops and conferences, such as ACM SAC, IEEE NCA, IFIPSEC or ARES. He is also a Senior Editor for IEEE Access.

\end{IEEEbiography}
\vfill

\end{document}